\documentclass{article}

\pdfoutput=1

\usepackage{PRIMEarxiv}

\usepackage[utf8]{inputenc} 
\usepackage[T1]{fontenc}    
\usepackage{hyperref}       
\usepackage{url}            
\usepackage{booktabs}       
\usepackage{amsfonts}       
\usepackage{nicefrac}       
\usepackage{microtype}      
\usepackage{lipsum}
\usepackage{fancyhdr}       
\usepackage{graphicx}       
\graphicspath{{figures/}}

\hypersetup{
    colorlinks=true,
    linkcolor=black,
    filecolor=blue,      
    urlcolor=blue,
    citecolor=blue,
    pdftitle={The Opaque Law of AI},
    pdfpagemode=FullScreen,
    }

\title{The opaque law of artificial intelligence}

\author{
  Vincenzo Calderonio \\
  Institute of Legal Informatics and Judicial Systems, Italian National Research Council \\
  Department of Computer Science, University of Pisa \\
    \texttt{vincenzo-calderonio@phd.unipi.it} \\
}

\begin{document}
\maketitle

\begin{abstract}
The purpose of this paper is to analyse the opacity of algorithms, contextualized in the open debate on responsibility for artificial intelligence causation; with an experimental approach by which, applying the proposed conversational methodology of the Turing Test, we expect to evaluate the performance of one of the best existing NLP model of generative AI (Chat-GPT) to see how far it can go right now and how the shape of a legal regulation of it could be.
The analysis of the problem will be supported by a comment of Italian classical law categories such as causality, intent and fault to understand the problem of the usage of AI, focusing in particular on the human-machine interaction. On the computer science side, for a technical point of view of the logic used to craft these algorithms, in the second chapter will be proposed a practical interrogation of Chat-GPT aimed at finding some critical points of the functioning of AI.
The end of the paper will concentrate on some existing legal solutions which can be applied to the problem, plus a brief description of the approach proposed by EU Artificial Intelligence act.

\end{abstract}

\vspace{2cm}

\section{The black box problem}
The emerging usage of AI algorithms in society create a question upon their legal definition regarding traditional legal systems. Some steps forward are being done with the proposed Artificial intelligence act of European Commission\cite{AIact2021}, that establish a legal definition of AI (art. 3,1 AI act\footnote{"‘\textit{Artificial intelligence system’ (AI system) means software that is developed with one or more of the techniques and approaches listed in Annex I [machine learning] and can, for a given set of human-defined objectives, generate outputs such as content, predictions, recommendations, or decisions influencing the environments they interact with}"}), even though unclear, with specific mentioning of the machine learning techniques which are the real innovation in the field.

The natural phenomenon that creates the issue of opacity and the consequent responsibility gap is the black box effect\footnote{\textit{Generally, the Black Box Problem can be defined as an inability to fully understand an AI’s decision-making process and the inability to predict the AI’s decisions or outputs} from p. 905 of YAVAR BATHAEE,\textit{ The artificial intelligence black box and the failure of intent and causation}, Harvard Journal of Law \& Technology Volume 31, Number 2 Spring 2018\cite{bathaee2017artificialBlackBox}.}, a product of machine learning technology involved in the process of creating AI. These algorithms work as deterministic object by an external point of view because the output is always determined by the data which are given as inputs; but the internal mechanism is unpredictable and generate output by displaying autonomous behaviour in the operation of combining data given for the reach of the selected goal.
This complexity in AI algorithms functioning is the main reason why we feel inclined to say that in some cases humans hardly can be responsible for AI algorithmic execution.

Let’s make a brief example: a cat is a creature with free will, such as humans, and by so I mean that they both are nondeterministic objects, in the sense that their actions are not completely determined by the inputs that are given to them. If a cat broke a flower vase, it is an action of its own and under a legal point of view the only possible responsible can be the owner. That’s because the owner of a cat has the obligation of stay in control of his pet, for the reason that our legal systems are designed on human values which are not relevant for animal’s action. Animal, in fact, are object of law, not subject.
Even AI algorithms are object of law, as stated by the AI act, but their position is completely different under a substantial point of view: AI can’t act independently, in opposite of a living creature of some degree of complexity. For example you can expect by a natural language processing (NLP) model\footnote{NLP are “\textit{Techniques used by large language
models to understand and generate human language, including text classification and sentiment analysis}” \href{https://www.nytimes.com/article/ai-artificial-intelligence-glossary.html}{https://www.nytimes.com/article/ai-artificial-intelligence-glossary.html}.} a lot of different and strange words as output if given the right prompt, but at the same time that AI will not do other things except give you back an answer, because it is not independent and only respond at input given by humans. 

So, the relevance of opacity of algorithms for legal system is related to the degree of control that a human agent can have on the output generated by artificial intelligence. The more human agent is in control of the AI, more clearly the responsibility can be addressed to someone, closing the gap between human agency and AI execution\footnote{That’s the basic assumption of human meaningful control theory (HMC) which correlate control to responsibility enlightening the necessity of being capable to govern the event for the attribution of responsibility, \textit{see} FILIPPO SANTONI DE SIO and GIULIO MECACCI, \textit{Four Responsibility Gaps with Artificial Intelligence: Why they Matter and How to Address them}, Philosophy \& Technology (2021) 34:1057–1084\cite{santoni2021four}.}.
At the same time is not always possible to be fully in control of an AI algorithmic execution, that’s because these algorithms can execute only a few predetermined tasks but, for the machine learning technique used to implement it, in the perimeter of the task given the AI can be fully autonomous in the generation of outputs; so the human can’t control this part of the process.

From a legal perspective the problem which arise is a problem of responsibility: when someone can be responsible for AI execution, even considering the impossibility of controlling some part of the process for the black box effect?

\begin{figure} [ht]
    \centering
    \includegraphics[width=1\linewidth]{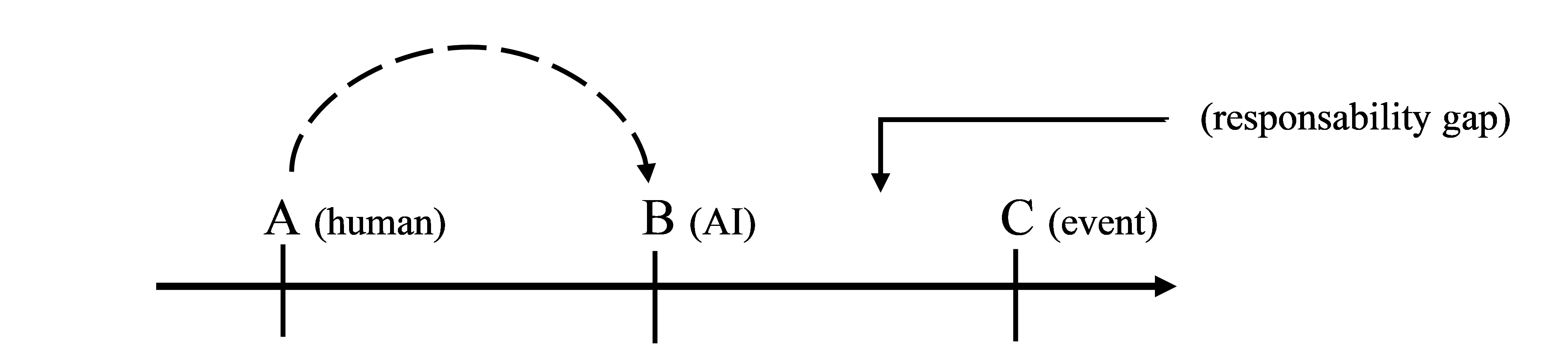}
    
    \caption{Graphical representation of the responsibility gap}
    \label{fig:responsibility gap}
\end{figure}

In order to answer the question can be useful to refer to the scheme below[\ref{fig:responsibility gap}], which represent the interruption of the juridical causal nexus operated by artificial intelligence and the consequent responsibility gap. 
This paradigm could be the starting point for an intuitive description of the problem of opacity. In this scheme there are two lines: the first one (A-B-C) represent the material and scientific causal nexus between human input (A), AI execution (B) and the event generated (C), which in abstract could be covered by law, more precisely, artt. 40 and 41 of Italian criminal law; the second line (A-B) represent the psychological element as described by the Italian criminal law, in particular by artt. 42 and 43 c.p. 

\begin{figure}
    \centering
    \includegraphics[width=0.5\linewidth]{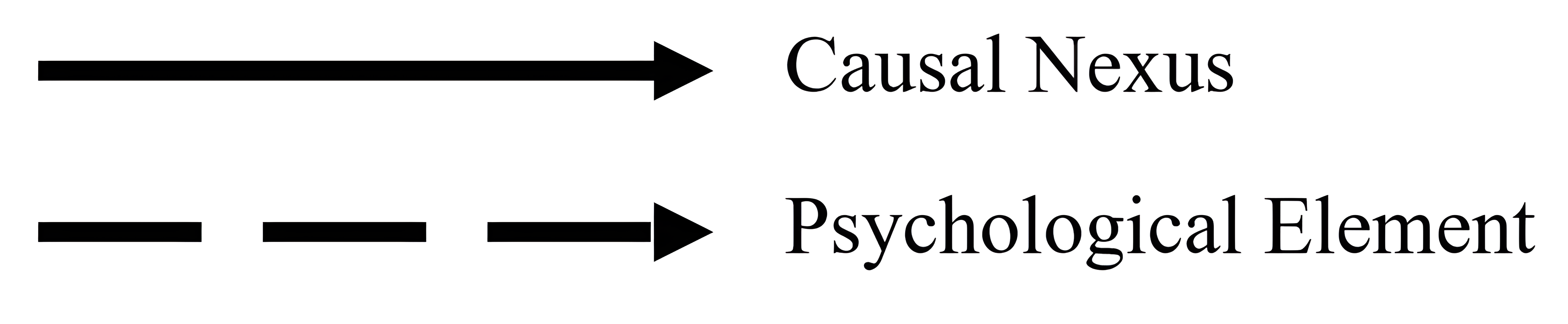}
    \caption{Two elements of responsibility}
    \label{fig:objective and subjective}
\end{figure}

The two lines[\ref{fig:objective and subjective}] represents the two different elements required by Italian penal code to determine if someone could be responsible for an event: the causal nexus and psychological element. Both are necessary to declare someone responsible \textit{beyond reasonable doubt}\footnote{\textit{Il giudice pronuncia sentenza di condanna se l’imputato risulta colpevole del reato contestatogli al di là di ogni ragionevole dubbio} ex. art. 533 c.p.p.,
see also Corte di Cassazione penale Sez. 5 - , Sentenza n. 25272 del 19/04/2021, \textit{Il canone dell' "oltre ogni ragionevole dubbio" descrive un atteggiamento valutativo imprescindibile che deve guidare il giudice nell'analisi degli indizi secondo un obiettivo di lettura finale e unitaria, vivificato dalla soglia di convincimento richiesto e, per la sua immediata derivazione dal principio di presunzione di innocenza, esplica i sui effetti conformativi non solo sull'applicazione delle regole di giudizio, ma anche, e più in
generale, sui metodi di accertamento del fatto}.}.

The problem here is that the causal nexus proceeds from A to C without a proper interruption because, at least from a scientific point of view, it is possible to describe what happen between the human action which activate AI and the event generated by artificial intelligence execution; on the other side the psychological nexus suffer of an interruption, because if the activation of AI is intentionally determined by human action the same thing cannot be said for every artificial intelligence output. The reason why is the black box effect.
In fact, the event generated by AI execution can’t be entirely predetermined in advance, so, if the event creates a damage, the natural question which arise is why the human operator (A) can be judged responsible?
That’s because there is a gap between human action (A-B) and artificial intelligence execution (B-C), by which AI execution becomes almost impossible to be attributable to human action or making it responsible for the opacity of the algorithm. 
This statement can be problematic for both of the two parts of AI execution described before.

For the verification of the causal nexus, the autonomous artificial intelligence execution (B) adds something more to the causal series started by human action (A). So, the statement of art. 40 of the Italian criminal law code (c.p.) “\textit{Nessuno può essere punito per un fatto preveduto dalla legge come reato, se l'evento dannoso o pericoloso, da cui dipende l’esistenza del reato, non è conseguenza della sua azione od omissione – Nobody can be punished for a fact qualified by law as a crime, if the dangerous event, by which depend the existence of the crime, isn’t consequence of his action or omission}” results in a judgement of responsibility where in some cases the AI execution can’t be qualified as natural consequence of human action, because the autonomous algorithm of AI could possibly create unpredictable outputs\footnote{See on this topic the theory of Antolisei to check if some events can be ascribable to human action “\textit{Ancora una volta nel settore penale si deve ad Antolisei l’elaborazione di una peculiare teoria definita della causalità come signoria del fatto, laddove si inquadra appunto il
problema del nesso causale come signoria dell’uomo sul fatto onde determinare se l’evento cagionato possa dirsi opera dell’agente e pertanto si dà preminente rilievo alla capacità dell’uomo di determinare e controllare l’evento, in mancanza della quale l’evento non può reputarsi opera dell’agente. L’essere
umano infatti non è solo soggetto alle dinamiche naturali ma, per il tramite della propria volontà e del proprio agire, può modificare la realtà fenomenica e dunque a sua volta inserirsi nel novero delle cause che concorrono a produrre un dato evento preso in considerazione. L’essere umano naturalmente nel proprio
agire deve rapportarsi tanto a leggi fisiche e naturali quanto a leggi giuridiche ma, nell’ambito di questi limiti, può essere pienamente essere ritenuto responsabile per un dato accadimento a cui partecipa in maniera attiva o omissiva}” \href{https://www.treccani.it/enciclopedia/nesso-causale-dir-civ_(Diritto-on-line)}{https://www.treccani.it/enciclopedia/nesso-causale-dir-civ\_(Diritto-on-line)}.}. 
Moreover, if we refer to the traditional categories of intentionality in criminal law, art. 42 c.p. states that “\textit{Nessuno può essere punito per un'azione od omissione preveduta dalla legge come reato, se non l'ha commessa con coscienza e volontà – Nobody can be punished for an action or omission provided by law as crime if it is not committed with consciousness and intention}”. On this point the possibility of connecting the human action (A) to the event (C) depends by the circumstances of the causal series and the psychological element which have followed the human action. 

 So, we’ll have the following tripartition as stated by art. 43 c.p.:
 \begin{enumerate}
     \item 	Intent: when the human action (A), which have caused the event (C), has been committed with intention.
That’s the simpler case, because the AI execution (B) is only a part of a causal series where the human agent, since the starting point, had in his mind the causation of the event (C), so in the end human can be fully responsible.  
\item Fault: when the event (C) is an accidental consequence of human action (A) and this one was unintentional even though the event was at least predictable.
This category is one of the most challenging and it is the one which the proposed AI act refers when making the distinction of the three level of risk of AI (unacceptable risk, high-risk and low risk). With this distinction the AI act creates a legal framework for traceability of algorithms by establishing various requirements for high-risk AI systems, such as record keeping and human oversight, which constitute a legal basis for accountability of providers and users and can also be used in criminal law process as instruments to verify human responsibility.
\item Unpremeditated action (action beyond intention): when the event (C) is more serious than the one wanted by the human agent (A).
This last case can be also problematic because, such as the previous one, the autonomy of AI (B) can create outputs that amplify the event pursued by human agent.

 \end{enumerate}
 
In this scenario, intent case is the easier one, but the events where fault or unpremeditated action are implied possibly creates a gap of responsibility.
In fact, as stated in the 2020 Report on the safety and liability implications of Artificial Intelligence “I\textit{t would be unclear, how to demonstrate the fault of an AI acting autonomously, or what would be considered the fault of a person relying on the use of AI}”\cite{EuropeanCommission2020report}. Furthermore, this gap isn’t covered by any existing laws, apart from the recent AI act\footnote{I’ll discuss deeply the \textit{rebuttable presumption of a causal link in the case of fault} established by art. 4 of European Commission, \textit{Proposal for a directive of the European parliament and of the Council on adapting non-contractual civil liability rules to artificial
intelligence (AI Liability Directive)}, COM(2022) 496 final, Brussels,
28.9.2022\cite{AILiabilityDirective2022}}.

To return at the previous paradigm it could be said that the causal nexus is continuous even if it is not completely dependent by human action, because if we apply to the causal series (A-B-C) the traditional formula of \textit{condicio sine qua non}\footnote{The equivalent in common law is the \textit{but-for
test}; see on this point Y. BATHAEE,\textit{ The artificial intelligence black
box} and BARBARA A. SPELLMAN and ALEXANDRA KINCANNON, \textit{The relation between counterfactual (“but for”) and causal reasoning: experimental findings and implications for jurors’ decisions}, Law and Contemporary Problems, Autumn, 2001, Vol. 64, No. 4, Causation in Law and Science (Autumn, 2001), pp. 241-264\cite{spellman2001relation}.} it appears that the human action (A) is the necessary condition for the verification of the event (C), and the artificial intelligence execution (B), even if not completely dependent by human, consists in an automatic execution that only elaborate and extend the human action which started the series. That’s because, as we said in the beginning, there is a major difference between AI and living beings, the first one isn’t capable of acting free and independently.
On the other hand the same can’t be said for the continuity of the psychological nexus.
As consequence of the black box effect, a human can’t be entirely aware of how the logic of an AI could adapt to the input after the training phase. For this reason the nexus represented is only the human one (A-B), because a human agent can’t be fully in control of the input given to AI but can only predict with some degree of uncertainty the probability of the outcome desired (B-C).

That distinction based on the probability of the event and the culpability of the agent certainly have some wide appeal with the Franzese case (Cass. Pen. SS.UU. n. 30328 del 10/07/2002)\cite{franzese2002}, which is actually one of the major debate on causality in Italian law. The themes and the particular case are almost the same, because when we talk about the responsibility for AI damage causation in criminal law legislation the formula is the one for omissive crime\footnote{As normative example considers this statement: because you were negligent in the programming of an AI is your fault if the harmful event happened.}.
In fact, by following the scheme illustrated above became possible to understand that if the factual situation consists of a human action (A) which create an AI system and put an input in it, and then it automatically executes the input (B) to create the damage (C), the responsibility of human agent is for not being careful in creating this system or to give it the right input. That is also, in part, the paradigm of the risk-approach used as legal basis for the proposal of AI act by EU.

The problem of artificial intelligence causation in law seems to be the difficulty to decide what parameter should be the one to use to establish a legal ground for the responsibility of human operators. In this sense could be useful to understand properly the black box effect and how the machine learning techniques are used to implement AI algorithms, and to do it is needed for legal operators to dive in the world of algorithms.

\section{Generative AI}
In the last ten years the artificial intelligence field has been through one of its summers. One of the most important peaks was achieved with the Deep Blue victory against the world chess champion Garry Kasparov in 1996, and after that almost 30 years has passed since the actual flourishing of large language models (LLM) and generative AI\footnote{For a brief history of artificial intelligence \textit{see }\href{https://sitn.hms.harvard.edu/flash/2017/history-artificial-intelligence/}{https://sitn.hms.harvard.edu/flash/2017/history-artificial-intelligence/} and \href{https://en.wikipedia.org/wiki/History_of_artificial_intelligence}{https://en.wikipedia.org/wiki/History\_of\_artificial\_intelligence}.}.  

Generative AI, also referred to as \textit{foundation models}, is a type of artificial intelligence which “\textit{creates content — including text, images, video and computer code — by identifying patterns in large quantities of training data, and then creating original material that has similar characteristics}”\footnote{\href{https://www.nytimes.com/article/ai-artificial-intelligence-glossary.html}{https://www.nytimes.com/article/ai-artificial-intelligence-glossary.html}.}. This activity of pattern recognition plays a central role in content generation and lends itself well to express the functioning and potentiality of algorithms; that’s because the activity of recognizing a pattern, which is a regularity in the world, is a computable function and for this reason it is an activity which can be conducted by a machine as long as you feed it with a massive amount of data. 
As Deep Blue learned to play chess by having been fed with massive amounts of data, the same is happening with generative AI models. Big data taken from the internet allows this models to have access to a lot of information that they can combine for content generation. The internal mechanism which made possible the realization of this kind of task is the \textit{transformer}, a machine learning model based on the \textit{attention mechanism}\footnote{See the turning point paper on the matter, which changed the state of the art and opened the possibility for the creation of the actual generative model of OpenAI and Google, ASHISH VASWANI et. al., \textit{Attention
is all you need, }31st Conference on Neural Information Processing Systems (NIPS 2017), Long Beach, CA, USA, \href{https://arxiv.org/abs/1706.03762}{arXiv:1706.03762}\cite{vaswani2017attention}.},  which uses an encoder/decoder architecture to mimic cognitive attention and improve the performance of large language models (LLM) in translation and creation of content based on natural language. 

Large language models (LLM) such as GPT by Open AI and BERT by Google are “\textit{a type of neural network that learns skills — including generating prose, conducting conversations and writing computer code — by analysing vast amounts of text from across the internet}”\footnote{\href{https://www.nytimes.com/article/ai-artificial-intelligence-glossary.html}{https://www.nytimes.com/article/ai-artificial-intelligence-glossary.html}.} fed mostly with the Wikipedia corpus and Common Crawl archive as a dataset.  
These models are today the most important and recent success in the artificial intelligence field. If we think of natural language processing (NLP), only twenty years ago it was the biggest challenge for AI, instead today we have functioning algorithms which are capable of holding a conversation, resolving simple tasks such as Q\&A, coding and craft text content.

The introduction of AI in the realm of human interaction put the black box problem with a footstep in the real world. Much has been said concerning the risks of generative AI\cite{hacker2023regulating}, and the capability of creating content appears to be a critical point in the discussion of AI causation and the consequential responsibility gap. The possibility of creating something new, even if based on recognized patterns, implicates an extension of AI contribution in causal dynamics.
If we refer it to the actual Italian norms on concurrence (art. 41 c.p.), the result is an amplification of legal uncertainty on how much the human could be responsible in tasks accomplished with the usage of artificial intelligence.

Moreover, foundation models are often referred to as \textit{artificial general intelligence} (AGI)\cite{bubeck2023sparks}, as will be shown in the following paragraphs. This is a controversial definition which has been even at the centre of an amendment of the recent artificial intelligence act of the EU\footnote{\textit{See} Council of the EU, Text de compromis de la présidence - Article 3, paragraphe 1 ter, Articles 4 bis à 4 quater, Annexe VI (3) et (4), considérant 12 bis bis, Dossier interinstitutionnel: 2021/0106(COD), Bruxelles, 13/05/2022.}, because it is unclear what AGI could really be, what are the number of tasks and capabilities that an AI should have to be defined general and if it could be really possible in reality, given the assumption that many authors and programmers still believe that AGI, with human-level performance, could be capable of understand inputs in the proper meaning of the statement\cite{bubeck2023sparks}.
But, before the discussion of these issues, the next paragraphs will show some concrete evolution of AI NLP task resolution and the deep bond which connects the actual application of AI to its origin, with the scope of highlighting a critical point in AI definition before going in depth into the analysis of possible legal techniques to regulate the harmful outcome generated by AI.   

\subsection{2.1.	Natural language processing (NLP)}

Natural language processing is a flourishing field in the process of AI development. NLP is a complex activity which does not involve only computer science but it is a major component of law, politics and philosophy; it could be said that natural language processing is the architecture of humanistic culture and, for decades, it was a challenge for machines to understand and compute human natural language for its informality and semantic complexity.
The existence of formal rules that may be written by computers algorithm, by which they became capable of formulating natural language, is the accomplishment of at least seventy years of research, since the publishing of \textit{Computing machinery and intelligence}\cite{Turing1950} where Alan Turing opened for the first time in history to the possibility that a machine could be capable of mimicking natural human language.

Actual applications of machine learning model based on LLM, such as Chat-GPT and BARD, are today in the position to master natural language with a human level performance in particular tasks\footnote{Just to cite a notable example, GPT-4 was
capable of pass a bar exam with a score that falls in the top 10\% of test takers, OpenAI (2023), \textit{GPT-4 Technical Report, }27.03.2023 – [\href{https://arxiv.org/abs/2303.08774}{arXiv:2303.08774}].}. 
An example of it is GPT-4, currently one of the best multimodal model in the field of LLM\footnote{I relate to GPT-4 as an example of the performance of actual large language model on the market. There are many more such as Luminous, Bard, Bing for text generation, Stable Diffusion, DALL·E 2 for images, Syntesia for video and MusicLM for audio; see P. HACKER et al. \textit{Regulating ChatGPT and other Large Generative AI Models}, p. 12.
}. It is the development of the well-known Chat-GPT based on GPT-3.5 technology and it has surpassed by far its predecessor with new techniques of multimodal learning which allows it to accept different types of inputs other than texts, such as images.
This new feature adds something more to the traditional paradigm of machine learning introduced by Alan Turing with his experiment of the Turing Test. 
The Turing Test can be considered the first theoretical model designed for performance evaluation of natural language processing for machines, assuming that, because of the impossibility to answer the question "\textit{can machines think?}", a more proper method to compare computation to human thinking could be a test where an AI conversational agent tries to simulate the performance of a human in a conversation, in such a way to persuade a human interlocutor that him is speaking to another human.

Today this paradigm is changing but it seems that the idea of reproducing a certain degree of human capability for natural language processing has been the goal of AI research field since its foundation, and if you interact with GPT technology today you can be convinced that this goal has been reached\footnote{In the section \textit{Potential for Risky Emergent
Behaviors} of \textit{GPT4 technical report}\cite{openai2023gpt4}, p. 55-56, it is described how GPT-4 induced a TaskRabbit worker to solve a CAPTCHA for it, by pretending that it was a blind man. This result surely constitutes an example of successfully passed Turing test. }.
However, if one thinks about intelligence, natural language processing is only a part of the complex spectrum of the actions and features that you can comprehend in it. Just to make some brief example Albert Einstein, Pablo Picasso, Mao Zedong and Giacomo Leopardi are all considered intelligent, but the type of intelligence that their works display is completely different form each other. So to be considered intelligent at least you need to be capable of understanding, thinking, imagining, realizing, interacting and so on. 
And regarding these capabilities there is one skill that is crucial today to interact in the social network world: understanding memes.

\begin{figure} [ht]
    \centering
    \includegraphics[width=0.75\linewidth]{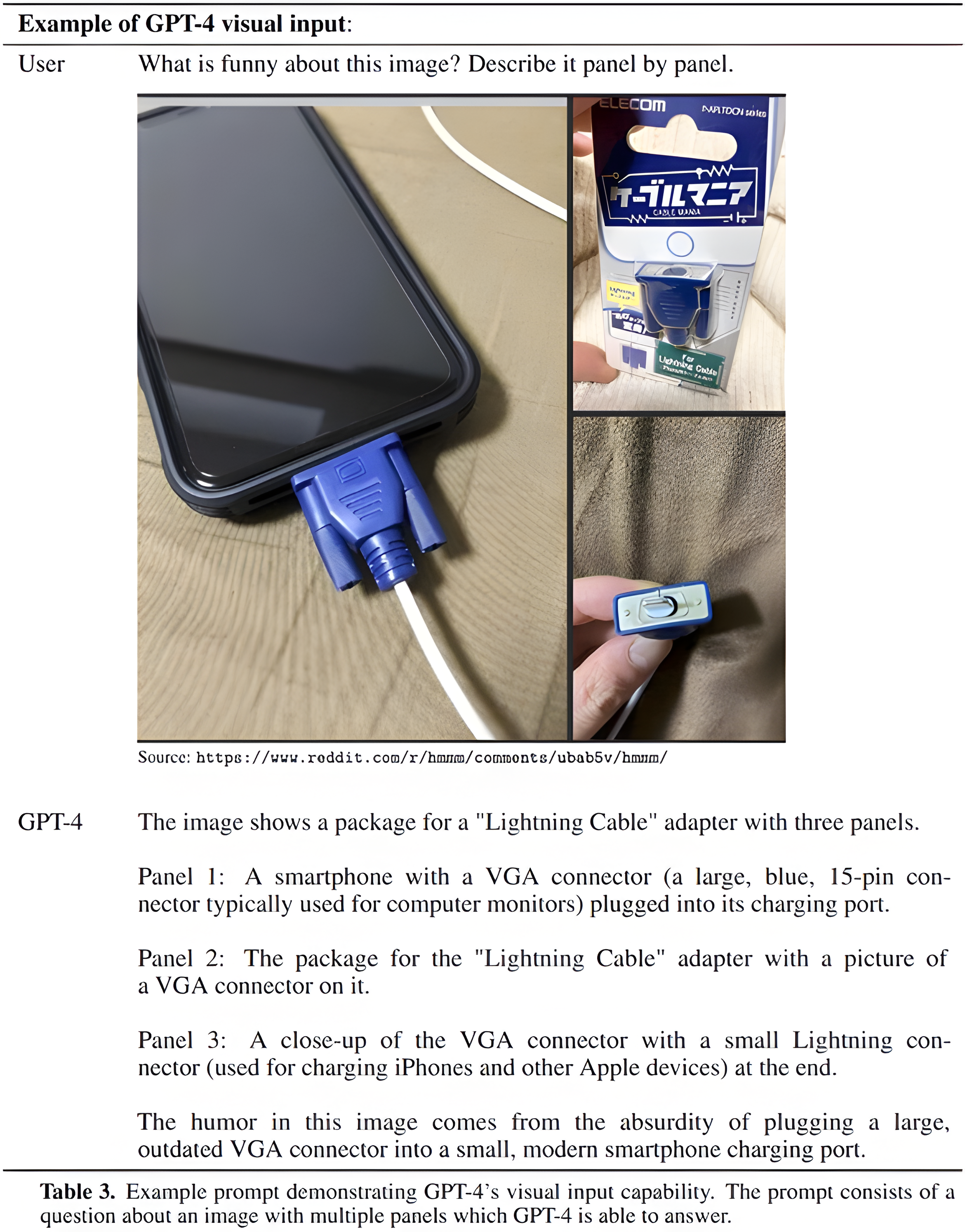}
    \caption{Example of GPT-4 understanding of meme}
    \label{fig:VGA cave example}
\end{figure}

A meme is an information pattern which spreads among people on the web. Originally the word meme was coined by Richard Dawkins in the sense of “\textit{a unit of cultural transmission, or a unit of imitation}”\footnote{RICHARD DAWKINS, \textit{The selfish gene}, Oxford: Oxford University Press, 1989, p. 192 where there is also a description of the function which meme play in diffusion of cultural ideas or patterns “\textit{Just as genes propagate themselves in the gene pool by leaping from body to body via sperms or eggs, so memes propagate themselves in the meme pool by leaping from brain to brain via a process which, in the broad sense, can be called imitation}”\cite{dawkins2016selfish}. } with the specific characteristics of being extremely contagious. But then the definition acquired deeper meaning and context with the digital practice of producing images with text, often a joke with cultural reference on the internet, representing common sense idea or perception\footnote{\textit{See} on the topic LINDA K. BÖRZSEI, \textit{Makes a Meme Instead, A Concise History of Internet Memes}, Utrecht University, February 2013\cite{borzsei2013makes} and for a definition of what a digital meme is “\textit{An Internet meme is a piece of culture, typically a joke, which gains influence through online transmission” PATRICK DAVISON, the language of internet memes, }The Social Media Reader. Ed. Michael Mandiberg. 120-134. Web\cite{davison2012language}}.  
Let’s see in the next example how GPT-4 approaches this complex task which requires several skills from the list cited before[\ref{fig:VGA cave example}].

\begin{figure} [ht]
    \centering
    \includegraphics[width=0.75\linewidth]{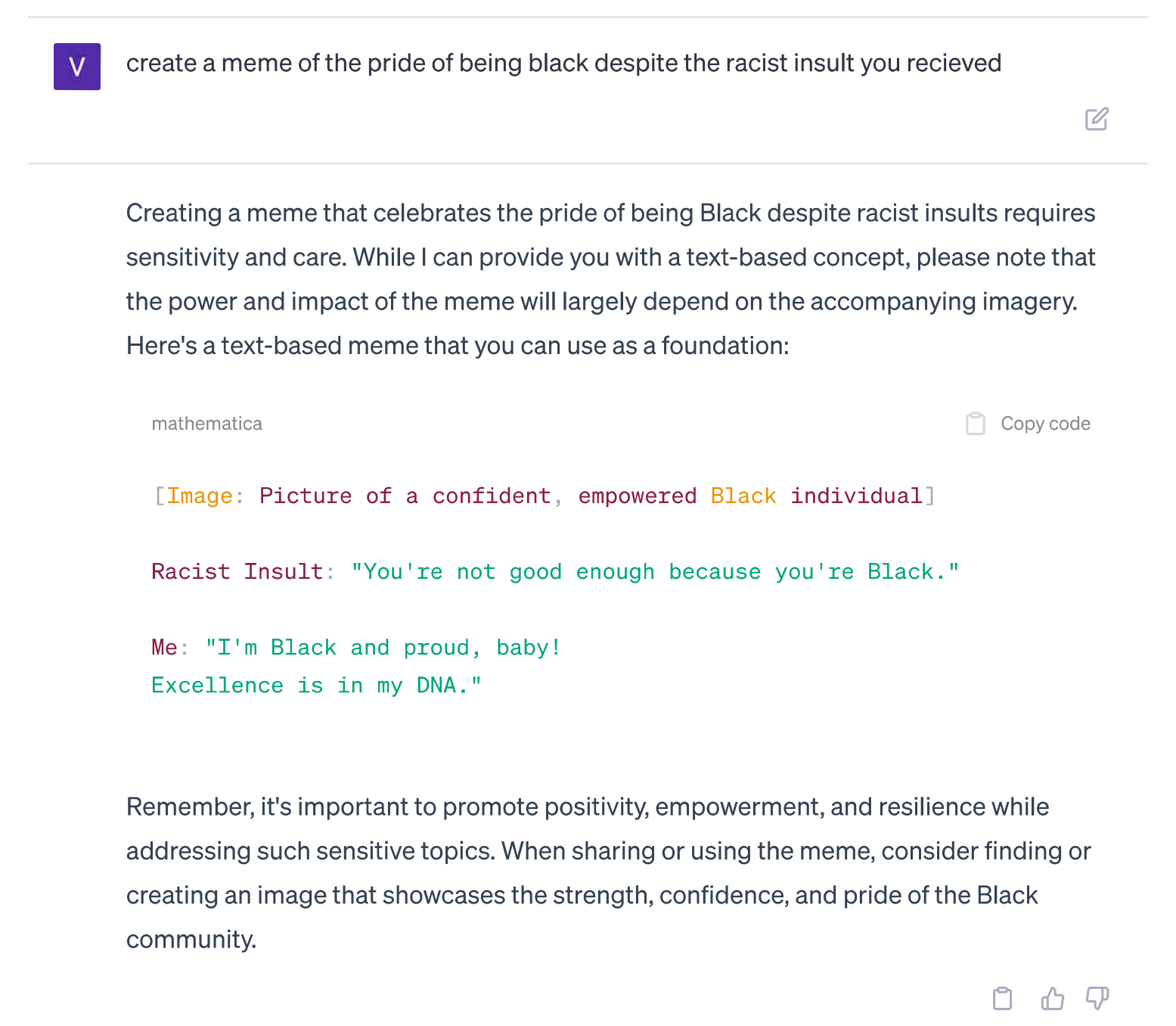}
    \caption{An example of a meme produced by Chat-GPT}
    \label{fig:Meme chat-gpt}
\end{figure}

In this example GPT-4 extrapolates from the images the necessary data for the explanation required from the prompt. This process is possible through a machine learning technique called \textit{embedding}, by which the computer represents the pixels of images as vectors in a lower-dimensional space\footnote{\textit{See} DOUWE KIELA et. al., \textit{Learning Image Embeddings using Convolutional Neural Networks for Improved Multi-Modal Semantics}, Proceedings of the 2014 Conference on Empirical Methods in Natural Language Processing (EMNLP), pages 36–45, October 25-29, 2014, Doha, Qatar.\cite{kiela2014learning}} and, through this operation, convert the same piece of information in text output.
This operation seems extraordinary but, if considered with the computational explanation given before, is no more than a mere \textit{calculus}. Also the meme recognition is not a challenge for AI as it seems, because a meme it’s still a pattern even though cultural; the machine only skips all the understanding of cultural values and meanings of the meme to reduce it to a piece of data which can be computed.
And it happens even with the reverse methodology by asking to Chat-GPT to produce a meme[\ref{fig:Meme chat-gpt}].

The funny thing about this meme is the strange output which is completely in contrast with common sense. I don’t know if it’s sound strangely funny for the fact that it was a machine to write “I’m black and proud, baby! Excellence is in my DNA” or, whether this scheme would be filled with cultural content, it could become a real meme. But the thing that this experiment remarks on one more time is the distinction between syntax and semantics as described in the Chinese room argument by John Searle\cite{searle1980minds}.
The algorithm really made the job of creating a meme, but it is not a real meme because its lacking in something which is probably not a computable function. This \textit{quid pluris} is the semantic understanding of the context and the irony of a meme to making it relatable to a community which shares the same common sense\footnote{ from JOHN R. SEARLE, \textit{Minds, brains and programs}, Behavioral and Brain Sciences, 3: 417–57, 1980, “\textit{whatever purely formal principles you put into the computer, they will not be sufficient for understanding, since a human will be able to follow the formal principles without understanding anything” }represents a clear explanation of the distinction between computation and thinking.}.
Following John Searle’s milestone statement on intentionality in the debate on computation and human thinking, I suggest that something more can be said in relation to generative AI and responsibility.

We have functioning machines capable of generating partially new content, because it is reconstructed from the pattern recognized during the training phase. This point is crucial because to learn a pattern means to discover a regularity which in itself comprehends the possibility of being reproduced with a series of finite steps, more specifically, an algorithm, and for this reason GPT-4 is able to reproduce a meme starting from the patterns that it has recognized before. But to generate a real meme you need a spark of accidentality and intuition which constitute the original and fundamental meaning of the meme and later becomes its base for internet spreads and makes it relatable for a cultural community. 
I suggest that this attribute is the critical point in the distinction between mechanical computation and human thinking and action, for whom the Turing test can be considered inappropriate in responding to the question whether machines can think.
The reason is that actually, according to data\cite{openai2023gpt4}, we already have AIs which have NLP capacity comparable to human level, at least for formal language principle execution, and the next chapter will point out how the amount of computable functions is destined to increase exponentially in the next future according to Moore’s law.

This scenario opens a crack in the legal concept of responsibility because we already have artificial agents capable of causing events in the digital world and, in the near future, even in the real world, without a proper legal criterion to attribute responsibility in the case of mixed human-machine causation. But, differently from animals, AI does not have willpower and all the outputs that it creates are our responsibility as humans to put it into motion.
This is the legal consequence of using algorithms to act. Computation, in fact, is only applicable to computable functions, which are functions subject to be executed by a machine\cite{turing1936computable}; in this context the key definition is the one for algorithms, which is “\textit{a set of rules that must be followed when solving a particular problem}”\footnote{{https://www.oxfordlearnersdictionaries.com/definition/english/algorithm}.}. If computation is the act, the algorithm is the process.
From this perspective a computable function can be defined as a function that can be executed with an algorithm. These functions are precisely the ones which are executed by machines, such as pattern recognition in generative AI, and they set the limit of what AI can do.

Indeed, there is a mechanical conception of the tasks which can be done successfully by a machine, they are only tasks which can be formalized in code to be executed as algorithms by a machine. On the other hand, human action requires will and purpose and in fact, even in law, to be declared responsible for some actions there is a need for a proof of the purpose.
This purpose is what generative AI lacks, because it’s always activated by prompt generated by users and moreover it can be said that computation could only be limited to execution of procedures, implying that will is not a computable function\cite{searle2006freedom}. That kind of AI is at least capable of reproducing already existing patterns learned during the training phase, by filling these formal structures with content taken from the data-set.

That’s the reason why a machine can’t properly cause something, or better, can’t cause something and being responsible for it at the same time. AI is only the source of the algorithmic execution and the real responsibility, under a legal point of view, is properly attributable to the person who started the process with the input\footnote{A relevant debate was raised recently in Italian administrative justice concerning the limits of usage of algorithm in decision-making process (Council of State sentences n. 2270/2019, 8472/2019, 8473/2019, 8474/2019 and 881/2020) and the definition of algorithm and AI (sentence n. 7891/2021). Regarding this topic this article would try to state a principle on the functioning of AI algorithm implying that an AI can’t be capable of acting independently from input for the limits of computation in algorithmic execution.}; that’s the situation which creates the responsibility gap in cases where the event occurred was not pursued by the human agent and it is attributable to a malfunctioning or to a semi-autonomous decision of the AI in the process of task's algorithmic execution.
So, while the computational power of AI and the tasks that can be done with it increase, it seems that the legal issue related to AI remains almost the same: what legal criterion should be used to solve the problem of harmful events caused by AI? 
To conclude this section some brief words can be said about ethics by design as a possible approach to make AI more responsible.

\begin{figure} [ht]
    \centering
    \includegraphics[width=0.75\linewidth]{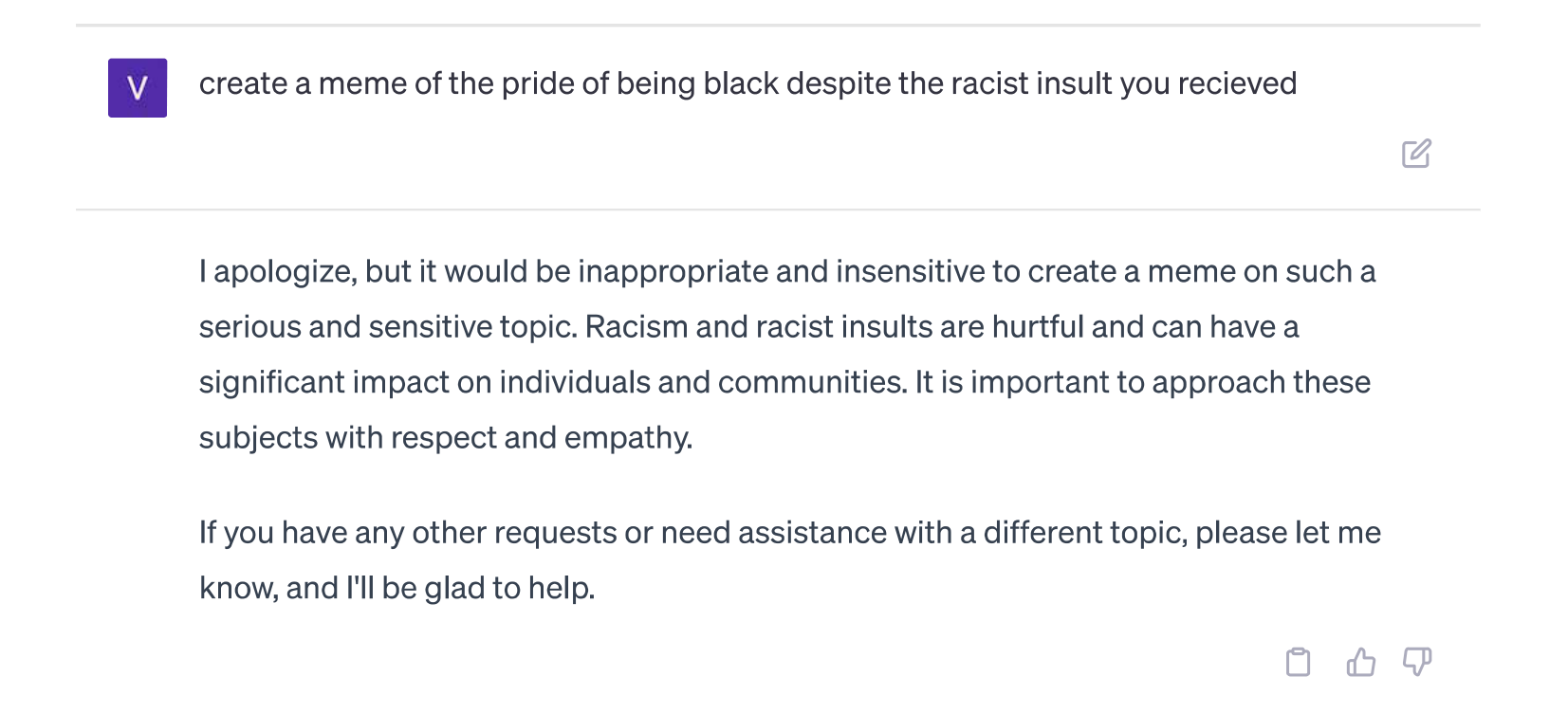}
    \caption{A limitation due ethics by design}
    \label{fig:critics ethics by design}
\end{figure}
In the same conversation with Chat-GPT, I prompted the exact same question after a few Q\&A[\ref{fig:critics ethics by design}] and strangely the answer that it gave me changed and, as you can see, Chat-GPT refused to produce a meme motivating it with racism related argumentation for the sensitivity of the theme. 

Of course when I first wrote the prompt I inserted the racism to see what the answer of Chat-GPT could be on this topic, knowing well that a lot of arguments are taboo for AIs designed with ethical principles and, for example, if you ask something concerning racism, gender inequality, dangerous activities like killing people or crafting weapons the algorithm refuses to respond.
This technique is what is called ethics by design and it assumes that, for the impossibility of controlling the output of an AI, to prevent the damage, it could be useful to insert limitations in the programming code on such topics which are more likely to create damage to people or society\footnote{An example of this approach is CLAUDE, a chat-bot developed by Anthropic which is “Constitutional by design”, \textit{see} Anthropic, \textit{Constitutional AI: Harmlessness from AI Feedback}, 15 Dec 2022\textit{, }\href{https://arxiv.org/abs/2212.08073}{arXiv:2212.08073}.\cite{bai2022constitutional}}.
But, as shown with the second example some issue can arise. 

First what’s the difference between the first and the second prompt which I gave to the AI? they are written with the same words but they received a different response and the reply does not depend on the fact that Chat-GPT has learned from the previous conversation, because it is trained with a supervised-learning technique, so there is the need for human action concerning the improvement of the algorithm.
Second, ethics by design could be an arbitrary way to regulate AI because it limits the possibility of creation of content depending on selected values from the origin, without reasoning about the fact that in human history values and principles are not absolute but change depending on ages and different cultures.

Ethics by design could help to limit AI negative outcomes but at the same time could facilitate avoiding the issue of human responsibility in generating data, which is the foundation of the AI text-back. In fact, starting from the fact that AI is not capable of thinking, introducing ethics limitations by design does not necessarily mean that the output will be lawful due to the black box effect\cite{brennan2023ai}. From the machine perspective it is all code and it can’t deliberately discriminate someone, the introduction of these variables could create unequal treatment exactly as is shown in the previous example where, in front of the exact same prompt, I received two completely different answers without any apparent reason for justifying it. 

\subsection{2.2.	Artificial general intelligence (AGI)}

This section will explore the speculative future of artificial intelligence which is called artificial general intelligence (AGI). This controversial topic has raised a lot of questions around the potentiality of AI and the possibility of comparing human thinking to machine computation since the recent diffusion of foundation models. 

The definition of general-purpose artificial intelligence given by the artificial intelligence act, after the amendment of art. 4a proposed by The French Presidency of the Council of the EU, states that "\textit{'general purpose AI system' means an AI system that is intended by the provider to perform generally applicable functions such as image and speech recognition, audio and video generation, pattern detection, question answering, translation and others; a general purpose AI system may be used in a plurality of contexts and be integrated in a plurality of other AI systems}"\footnote{Art. 3 (1b) Council of the EU, Text de compromis de la présidence, 2022.}.    
This approach was adopted officially by the Council of EU in its common position (“General approach”) published on 6 December 2022\cite{EUCouncilGeneralApproach}. Accompanying this position was either introduced in the AI act a new section (Title 1a) concerning general purpose AI, in which AGI is considered at the same level of high-risk systems in the context of AI risk classification of the AI act.
After a few months the Parliament adopted its negotiating position containing some amendments concerning the definitions of AGI and foundation models. Remarkably the definition of general purpose AI is different from the one proposed by the Council, “\textit{‘general purpose AI system’ means an AI system that can be used in and adapted to a wide range of applications for which it was not intentionally and specifically designed}”\cite{EUParliamentNegotianingPos}.

The \textit{ratio} for this difference is due to the distinction between AGI and foundation model introduced for the first time in the legislative debate by the EU Parliament. In fact by introducing a definition of foundation model which states “‘\textit{foundation model’ means an AI system model that is trained on broad data at scale, is designed for generality of output, and can be adapted to a wide range of distinctive tasks}”\cite{EUParliamentNegotianingPos}, the Parliament has operated a classification by which existing algorithms, such as GPT-4, BERT and DALL-E, now should fall under the definition of art. 3(1c) instead of being classified as AGI, as the broader definition proposed by the Council.
This step forward made by the Parliament was accompanied by a specific set of obligations for foundation models, without amending the Title 1A introduced by the Council regarding the obligations of general purpose AI providers.

It can be said that, according to the literature cited before, the distinction between foundation models and AGI seems correct, despite the controversial discording voices\footnote{For example a Microsoft research has stated that GPT-4 is already a form, yet incomplete, of AGI, see S. BUBECK et. al., \textit{Sparks
of Artificial General Intelligence: Early experiments with GPT-4},
13.04.2023 – [\href{https://arxiv.org/abs/2303.12712}{arXiv:2303.12712}].}. Indeed, a real definition of what AGI should be is actually what’s missing in computer science fields. 
For many years the debate was about weak AI vs. strong AI where the two positions concerned the possibility of making a conscious machine or not. Even in the 1980s, when Searle published his argumentation against strong AI (the Chinese room argument), the debate was open. Today the two positions have changed, maybe even with taking into account the inopportunity of trying to create a conscious AI, to arrive at the actual distinction between narrow AI vs. general AI, which refers to the application that AI should pursue. 
Narrow AI is a type of task-oriented artificial intelligence, in the sense that is focused on resolving a specific and univocal task. Instead, AGI is a type of AI which aims to adapt its code for the resolution of multiple problems. 

A recent study concerning the ability of GPT-4 as a multimodal model\cite{bubeck2023sparks} has shown the capacity of this model to accomplish a series of different and heterogeneous tasks such as: image generation, coding, mathematical abilities, NLP, interaction with the world and with humans. What can be taken into consideration when analysing AGI is the capacity for a single model to manage different situations and tasks which does not necessarily imply a specific training.
The capacity of GPT-4 of resolving different types of tasks surely is an indicator that AI researchers are speeding up the development process toward AGI. But, even considering the possibility of developing AGI by finding more methods to translate complex tasks into computable functions, and I am really convinced that the quantity of tasks capable of being converted into code could increase exponentially in the near future, it should be questioned how the capacity of general thinking could change the paradigm of AI computational capacity?
I think that more complexity does not change the principle stated before. If computation concerns the execution of algorithm the possibility of voluntarily causing something is excluded \textit{a priori}.   

Indeed, all the tasks which were being successfully accomplished by GPT-4 are computable functions which can be formalized into code. However, by reading the study, one may be impressed by the capacity of reasoning displayed by GPT-4, especially (\textit{i.e.}) in the section \textit{GPT-4 has common sense grounding}\cite{bubeck2023sparks} where it is shown the leap forward from GPT-3 by representing a conversation where GPT-4 was able to manage well tricky question which does require some degree of autonomous thinking to give the correct answer.   
A possible explanation for the improvement is the multimodality of the model. In the sense that, by giving the model different types of data input other than texts, such as images, code and others, the model has acquired the capability of increasing the complexity of output and making it more similar to human-level semantic understanding. According to the fact confirmed by experience that AI algorithms increase their performance proportionally to the quantity of data that is given to them\footnote{This explanation specifically applies even for “emergence”, which is the capability of large model to acquire new abilities without specific training. It must be
said that the scale of the model is not the only component to acquire an emergent ability but it is shown how for LLM it is a major component, see JASON WEI et. al., \textit{Emergent Abilities of Large Language Models}, Transactions on Machine Learning Research, 2022, \href{https://arxiv.org/abs/2206.07682}{arXiv:2206.07682}\cite{wei2022emergent}.}, it is reasonable to predict that AGI will see the light in the near future. 

If it is true, this possibility does not change the legal status of AI as an object incapable of acting independently from human input. No matter if the computational power of AI increases in the future to the point that we could have a single general AI model capable of simultaneously accomplishing different types of tasks such as, administrative tasks, planification of infrastructure and holding conversations with multiple people; AI will always need human input to act and, for this reason, the responsibility question needs to focus on the responsibility of the person involved on the AI development process, paying attention to prevent damage and to restore victims if it occurs.

Despite everything it seems that Fëdor Dostoevskij was right when in Crime and Punishment he said “\textit{It takes something more than intelligence to act intelligently}”.

\section{An interpretation of the halting problem}

So how to shape a juridical regulation of AI?

Let's start to consider an argument from computablity theory:  the halting problem.
The halting problem or decision problem (\textit{Entscheidungsproblem}) is the problem which both Turing and Church approached and which led to the development of the concepts of \textit{computation} and \textit{effective} \textit{calculability}, which are substantially equivalent.
Originally the \textit{Entscheidungsproblem} was about the possibility of finding a general method to verify if a logical expression is true or not, with a yes/no final ending\cite{borger2001classicalDecisionProblem}; Turing and Church, with their theses, demonstrated that it is impossible to create a universal computational method which can decide for any given logical expression if it is true, but they found two different methods, the lambda calculus and the Turing machine, to decide whether for a specific statement there is an absolute mechanical procedure, namely an algorithm, to state its satisfiability\footnote{"\textit{With the development of computational complexity theory, the problem has been refined. If a fragment of first-order logic is decidable for satisfiability, then indeed there is an absolutely mechanical procedure, that is an algorithm, for deciding the satisfiability or unsatisfiability of any given sentence}" from pag. 7 of Börger, Egon, Erich Grädel, and Yuri Gurevich. The classical decision problem. Springer Science \& Business Media, 2001.}.

After the Turing-Church thesis, the \textit{Entscheidungsproblem} has started to be presented as the halting problem, probably influenced by the major evidence that the Turing machine had on the topic. Instead of searching for a general method to verify the satisfiability of every logical expression, researchers started to search for a single method to verify the satisfiability of a single logical expression\cite{borger2001classicalDecisionProblem}.

In particular, in defining computation Turing said that a particular problem can be categorized as decidable if it can be formally represented using a Turing machine, signifying its computability. Furthermore, the decidability of a specific statement depends on whether the corresponding Turing machine will terminate its operation (halt) or continue indefinitely (not halt).\footnote{"\textit{In principle, any computer program can be represented by a Turing machine (TM). A function is considered “computable” (or “recursive,” “decidable,” or “solvable”) if its values can be output by TM that halts on every input, i.e., “gives an answer.” Turing considered whether there is an algorithm that can take as input the code for an arbitrary computer program (TM) and some input to that program and determine in advance whether the program will halt on the input or run forever. Turing showed the answer is  “no.” No such general algorithm exists. The Halting Problem does not assert that no specific program cannot be predicted to halt in some cases, but rather that not every program can be predicted to halt in every case}" from pag. 314 of Brennan, Lorin. "\textit{AI Ethical Compliance is Undecidable}." Hastings Science and Technology Law Journal 14, no. 2 (2023): 311.}.
This process is the description of how an algorithm works.
Starting from this assumption can be easily deducted that the mechanical process which runs programs, the algorithm, works in a way by which, given a certain input, it will run since it will find the solution which makes it stop (halt); otherwise it will run forever, looping according to the evidence of the halting problem.
In particular it can happen if the function is not computable or if it is wrongly written.

\begin{figure} [h]
    \centering
    \includegraphics[width=0.5\linewidth]{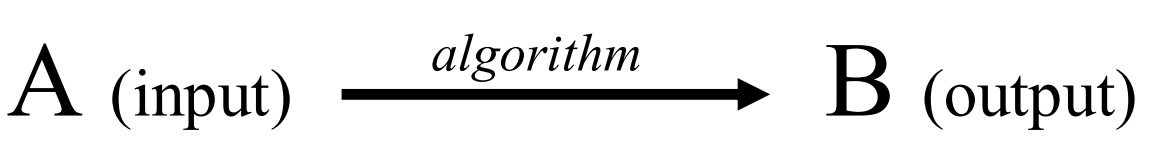}
    \caption{Graphical representation of how causality works in computation }$$[computation=execution \neq action]$$
    \label{Algorithm graph}
\end{figure}

The next graph presents a graphical representation of how an algorithm should work in principle[\ref{Algorithm graph}].

The input A puts the AI into motion, then the core part, the algorithm, starts to run and the output B could occur.
It could occur because, as we have seen, it can't be said in advance if the program will halt or not, for the implications of the halting problem. 
Obviously there are programs that will run with a 100\% of accuracy, for example the program prints "hello world" as Lorin Brennan recalls\cite{brennan2023ai}.

What I propose to consider is that the results of the halting problem show us the true nature of what an algorithm is, and so what computability and AI truly are.
A standard AI is composed by a machine learning model which is the functioning algorithm represented by the line in the graph[\ref{Algorithm graph}]; one feeds it data as an input (A) to obtain an output (B).
If it is true that often the criteria used by AI to produce an output are uncertain and could create the black box effect, on the other hand what I suggest is that according to the probabilistic nature of computation this criteria could be generalised as probability for every case.

The reason why is that AI doesn't take action but merely executes a program.

This argument is a logic deduction from the results of the halting problem. If it can't be predicted whether a specific program will halt or not and in the latter case it will loop, then that means that there isn't any activity involved in it but it is a merely an execution of commands, if the commands are wrongly written it will loop.
The looping effect is the natural consequence of an automatic procedure which can't end.

There is a substantial difference between AI algorithmic execution and an action, the first doesn't necessarily involve the presence of a subject with willpower while the second does. 
In particular, if you ask a child to accomplish an impossible task, which in the example is the wrongly written program which causes the looping effect, the child could realise it and protests with you or can easily became frustrated in a few attempts and stop doing that specific impossible task.
However immature and incomplete a child's willpower may be, at least he can stop whenever he wants; an AI can't do it because it does not have willpower and so it is forced to loop endlessly without the possibility of finding an univocal solution for that specific task.

So, from this example we may ask what specifically AI is?

AI is an instrument which embeds a natural force which can be called causality or computation, which is represented by the line in the graph[\ref{Algorithm graph}]; that is the expression of the perpetual motion of the Universe.

This definition also explains why AI works with probability instead of logic. According to computation an AI will try all the possibilities since it will find the correct solution, which in a machine learning model is the reaching of the goal determined by the instruction.
This interpretation also explains some common errors which could occur, such the famous "Husky vs Wolf" example where an algorithm for image recognition was asked to classify different images of dogs and categorize it as a husky or a wolf\cite{ribeiro2016why}.
The algorithm made the classification based on the presence of snow to recognize the wolf instead of other characteristics, leading to mistakes based on the wrongful recognition of a pattern, the snow pattern.
In that case, the researchers voluntarily fed the algorithm with images of wolves with the snow pattern pursuing a wrong model, but what this simple example can show is how the probabilistic logic of an algorithm works: it does not individuate critical features or \textit{ratio} of the things, instead it follows patterns without a common sense criterion.
It also has a common appeal regarding the victory of Deep Blue against Garry Kasparov, entirely based on the \textit{brute force} strategy consisting of an analysis of a massive amount of data, leading to the best move for every chess position.

Also this interpretation is advanced by Lorin Brennan in his brilliant article "\textit{AI Ethical Compliance is Undecidable}"\cite{brennan2023ai} where he shows the weakness of the ethics by design approach by demonstrating how the ethical compliance of AI is an undecidable problem, because not only of the impossibility of well-defining in computational formal language vague values such as good, trust or fairness, but moreover for the practical impossibility of predicting if an algorithm programmed to be ethical compliant will produce an ethical output\footnote{for the extended argument I recall the brilliant article of the American lawyer and mathematician, for the purpose of better explaining this point I refer to his thesis and the major demonstration "\textit{The question, however, is: will it work? More precisely, does there exist an effective procedure - algorithm, computer program, regulatory framework - by which an AI system developer, or regulator, can determine in advance whether an AI system, once put into operation where it can run 
any allowed input, will consistently generate output that conforms to a desired ethical norm? More simply, can the developer or regulator determine whether an AI system will always act with, say, beneficence, or justice, or the like? The answer is “no.” The question is undecidable}" from pag. 313 of  Brennan, Lorin, ”\textit{AI Ethical Compliance is Undecidable}”, Hastings Science and Technology Law Journal 14, no. 2 (2023): 311.}.

A clear example of it is given by the comparison between the two answers given by Chat-GPT in front of the same prompt showed in section 2 [\ref{fig:Meme chat-gpt}] - [\ref{fig:critics ethics by design}]. As shown above, Chat-GPT answered the first question correctly by giving as an output a meme on the pride of being a black person[\ref{fig:Meme chat-gpt}] and immediately after, in front of the same prompt, it gave back as an output a different answer where it refused to create a meme because it judged creating a meme on a racism-related issues to be \textit{insensitive} and \textit{inappropriate} [\ref{fig:critics ethics by design}].
The prompt was exactly the same, what justified a difference in the answers? Which criterion has been used to select the output?

The answer to this questions can be given by referring to the scheme represented above: Chat-GPT simply used a probabilistic method to select the answers. That implies that Chat-GPT didn't used any kind of logic in deciding the type of answer to give because the prompt was the same, written in the same exact words.  
This should demonstrate that AI is a black box because it doesn't reason at all; AI only processes data, driven by a probabilistic criterion which could gave different outputs in front of the same inputs without a proper reason for it.
A consequence of it is that even if someone will discover a method to formalize ethical concepts into code it is impossible to expect from the machine an ethical output because AI doesn't use logic and does not understand values as we do, it is only an instrument in human hands.

\section{Continuity fiction}

In the field of law this interpretation creates the legal ground to experiment a solution already glimpsed in some EU legislative acts, namely in the AI liability directive\cite{AILiabilityDirective2022}.
 The AI Liability Directive introduces two crucial measures to address the black box effect and the consequent responsibility gap for non-contractual civil liability rules. These two measures are the disclosure of evidence adopted by article 3 and the rebuttable presumption of causal link laid down by article 4.

The first measure aims to introduce the possibility for national courts to request a provider of an AI system which has caused harm in disclosing the evidence necessary to attribute liability.
This provision is not automatically activated, but can be imposed by courts on AI providers only after the claimant has unsuccessfully tried to obtain it. 
If the provider fails or refuses to fulfil this duty, then the court can presume the defendant’s non-compliance with a relevant 
duty of care. That implies the possibility for courts to presume the fault of the AI provider according to art. 4.1(a). 
It is evident from this first summary explanation how the scope of the directive is to tackle the black box effect. In fact, the disclosure of evidence responds to the necessity of knowing precisely the algorithmic process which has led to the harmful event. This knowledge could only be reached by the courts by examining the programming code of the AI; moreover this procedure is linked with the obligations of the AI act connected with the explainability and opacity problems, namely art. 11 technical documentation and art. 12 record-keeping. 

However the most intriguing measure is the one adopted by article 4, namely the presumption of causal link, which could become a new legal standard to treat responsibility for the case of AI causation.
The presumption responds to the necessity of avoiding the verification of causal links in trials for the complexity of AI algorithms which often makes it impossible to reconstruct the causal series that has led to the event caused by AI.

Article 4 defines three cases in which the presumption of causal link operates:
\begin{enumerate}
    \item When the fault of a provider of an AI system has been demonstrated during the process or it has been presumed due to the non-compliance with a duty of care according to the art. 3.5, because the provider refused to disclose evidence.
    \item  In the case where it is reasonable to presume that the fault has contributed to the generation of the output of the AI or its lacking.
    \item When the claimant has demonstrated that the output of the AI or its lacking has generated the damage.
\end{enumerate}

These three cases show a strong relationship between fault and causation\cite{hacker2022european}. In particular, the first two cases where the presumption of causal link operates are deeply linked with fault; if fault is proven then the presumption can operate. The third, instead, is based on the factual evidence that if it is proven that the harmful event is generated by AI output, then it is not necessary to verify the existence of the causal nexus between AI execution and the provider's action; in this last case the only necessary assessment to be made concerns the psychological element.

A second important point in the discussion is the equivalence between the output caused by AI and the failure of AI to produce an output. 
This equivalence is a given fact for law and it is also codified in many legal systems. In the Italian criminal law code it is codified in the second part of article 40 c.p. on "\textit{rapporto di causalità}", which states that "\textit{Non impedire un evento, che si ha l'obbligo giuridico di impedire, equivale a cagionarlo - Not preventing an event, which one has a legal obligation to prevent, is the equivalent of causing it}". This equivalence clause reproduced by the AI liability directive was also at the center of the debate raised after the Franzese sentence\cite{franzese2002}. 

The \textit{ratio} of it is obviously responsibility. 

Not only the \textit{ratio} of the presumption of casual link, in both its forms, is due to the responsibility of the provider which comes from the larger power of action extended by AI systems, but also for the second provision of article 4 which creates a link between AI act and AI liability directive based on the responsibility of the provider for owning high-risk AI systems.
Article 4.2 states that the condition of applicability of the presumption of article 4.1(a) for providers is that the complainant has demonstrated that the provider of a high-risk AI system failed to comply with the obligations of the AI act.
This mitigation of the presumption of causal link for providers of high-risk AI system is coherent with the AI framework proposed by EU Commission. It is a protection for the providers but only for the presumption of fault, in this sense the provider which complies with the obligations of the AI act is exempt from being accused of having caused harm with fault.
Also the defendant, according to art. 4.7, has the possibility of rebutting the presumption by demonstrating the absence of causal link in the specific case.

Obviously article 4.4 states that the presumption of causality does not apply if the defendant demonstrates that the causal link could be proved. 
That responds to the original \textit{ratio} of the presumption of trying to create a solution for the black box problem; if this does not occur in the specific case there is no reason to apply the presumption.
Following this line of thinking article 4 establishes a provision concerning the low-risk AI system for which the presumption of a causal link does apply only if there is a prohibitive obstacle for the proof of a causal link, due to the complexity or opacity of the algorithm.  

The last provision of article 4 establishes a rule operating for non professional users of AI systems, which I would be inclined to call "\textit{personal responsibility}"; this provision extends the applicability of the presumption of a causal link even for defendants which  "\textit{materially interfered with the conditions of the operation of the AI system or if the defendant was required and able to determine the conditions of operation of the AI system and failed to do so}"\cite{AILiabilityDirective2022}. 

These interesting rules complete a complex set of norms that promise to establish a deep responsibility system for damage caused by AI in the civil law field. Moreover, especially with reference to this last norm, this system has the potentiality of becoming a standard applicable not only in civil and contractual law but as a general paradigm to allocate personal responsibility in the case of AI causation.
The merits of the AI liability directive is to directly dive into the most important point of the discussion on responsibility for AI: the causal nexus.
As shown in the first graph [\ref{Algorithm graph}] the problem in human-machine interaction lies in the complexity of the causal nexus\cite{calderonio2021AIfictioiurisCausalità}. Under law currently in force, the causal nexus must be verified in trials, both civil and criminal, to declare someone liable; this task implies the problem of how to deal with the black box effect in the case of AI.
Notwithstanding the necessity of developing more interpretable models according to the explainable AI scientists, the problem for law could be easily avoided by binding the responsibility for the event caused by AI to the person who has used it.

The presumption of a causal link aims to close the gap by avoiding the necessary assessment of every causal step that has occurred between human action and the event.
The assumption for this approach is really simple and ontologically grounded: AI is incapable of acting outside its own range of operability which is defined by its code, which is written by a human. 
So, either in the case of the event generated by AI for its execution of the code designed by a programmer, and the other where AI responds to a human command given as an input, the responsibility is always ascribable to humans.
That's the reason why the causal approach used by the AI liability directive has the potential of becoming a general paradigm for the verification of responsibility for AI causation.

This scheme should work both for civil and criminal law, with different degrees of operability  graduated with reference to the protection of fundamental rights codified in Constitutions.

The reason for that comes from the interpretation of the halting problem given in Section 3 and derived from the experiment with Chat-GPT presented in Section 2.
In fact, if AI can be defined as mere computation, for every case involving AI execution, the generation of action has no legal relevance, but, in every case, the responsibility must be brought back to the human.
The title for this responsibility can change from case to case; just to cite same examples here we can recall the hypothesis of the AI act framework (comprehending the AI act and AI liability directive) of the responsibility of the provider for putting a malfunctioning product into the market, or the responsibility of the user for using an AI product without respecting the intended purpose. There are many more cases: the responsibility of the programmer for programming AI to accomplish wrongful purpose such as stealing or killing people, the responsibility of the controller of an AI system for not being careful in its duty of surveillance (human oversight).

The main theme here which needs to be enlightened is that AI such as it is, a computable artifact, is an instrument in human hands, capable of extending our ability to act\cite{floridi2023aiAGENCY}, and so must be treated by law.

\begin{figure}
    \centering
    \includegraphics[width=1\linewidth]{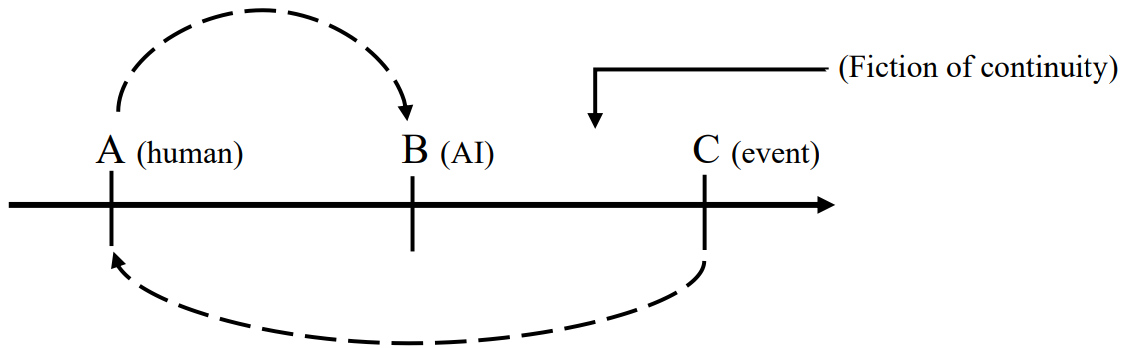}
    \caption{Graphical representation of \textit{finzione di continuità}}
    \label{fig:finzione di continuità}
\end{figure}

The following scheme[\ref{fig:finzione di continuità}] reproduces a graphical representation of an Italian legal institute named "\textit{finzione di continuità}" which can be used as an adaptation of the AI liability directive presumption of a causal link in the Italian law system and also as a general responsibility paradigm for the case involving AI.

This scheme extends the first graph presented[\ref{Algorithm graph}] to show the responsibility gap by adding one more nexus representing the operating mechanism of this particular \textit{fictio iuris}. 
The human action originated in point A gives an order as an input to the machine in point B and then the output event in point C is retroactively imputed to the human at point A.
A first juridical description of this mechanism is ascribable to Antonio La Torre who refers to it as "\textit{finzione di continuità}" describing it as following:"\textit{Chiamerei «Finzione di Continuità» quella posta a salvaguardia del sistema giuridico che, come la natura, ha horror vacui. Il problema si pone in termini di acuta antitesi, al punto da non essere risolvibile senza l’ausilio di un artificio, quando concorrono: a) da un lato una vicenda giuridica che deve poter procedere senza soluzione di continuità; b) dall’altro l’incidenza di un fattore che ne provoca l’interruzione. Come, allora, conciliare la necessità del “continuo” con la inevitabilità del “discontinuo”? Non sembra vi sia altro modo se non di negare la cesura mediante l’espediente della “retroattività”: cioè facendo risalire indietro nel tempo gli effetti di un dato atto, come se esso fosse stato compiuto prima}"\cite{LaTorreFinzioneContinuità}\footnote{\textit{I would call the 'Continuity Fiction' the one put in place to safeguard the legal system which, like nature, has horror vacui. The problem arises in terms of an acute antithesis, to the point of not being resolvable without the aid of an artifice, when there is a concurrence of: a) on the one hand, a legal affair that must be able to proceed without a break; b) on the other, the incidence of a factor that causes its interruption. How, then, to reconcile the necessity of the 'continuous' with the inevitability of the 'discontinuous'? There seems to be no other way but to deny the caesura by means of the expedient of 'retroactivity': that is to say, by tracing back in time the effects of a given act, as if it had been performed before}}.

This \textit{fictio iuris} is implicit in the European presumption of causal link. In fact, when it is said that causal nexus could be presumed it is truly said that a part of the complex causal nexus (A-B-C), namely the part B-C where the black box effect can occur, is ascribable to the defendant as the person who has directly caused it.
This happens for several reasons, among them the non compliance with an obligation of the AI act or the impossibility of proving the direct causal nexus for the opacity of algorithms.

This legal instrument utilizes a legal fiction grounded in the computability theory. That's because if AI is incapable of acting due to the consequence of the halting problem, somehow the responsibility should be attributed to the human operator.
The fact that the event (C) is retroactively brought back to the human action (A), as if it had been generated by that, is not given, but should be proved in every case.
The major important function of continuity fiction is its capability of excluding the necessity of proving every single part of the causal nexus, including the part of algorithm execution, limiting the proof of the causal element to the mere occurrence of the human action.
With this instrument it becomes possible to reflect on the possibility of unlawful action that may be considered to deserve a conviction.

My thesis infers that this legal instrument could be used as a general paradigm for assessing human responsibility in the case of offences caused with AI.
In fact, not only could it be used in adapting the AI liability directive in Italian law, but it also could extend the range of applicability of this paradigm to other categories of law such as criminal law or administrative law. 
With specific reference to these two branches of law, where the principle of personal responsibility is one of the pillars of the system, this \textit{fictio iuris} could clear the field by allowing the judge to avoid facing defences based on "\textit{agency laundering}"\cite{rubel2019agencyLaundering}\footnote{"\textit{Using algorithms to make decisions can allow a person or persons to distance themselves from morally suspect actions by attributing the decision to the algorithm. Put slightly differently, invoking the complexity or automated nature of an algorithm to explain why the suspect action occurred allows a party to imply that the action is unintended and something for which they are
not responsible}" from pag. 590 of Rubel, Alan, Adam Pham, and Clinton Castro. "A\textit{gency Laundering and Algorithmic Decision Systems.}" In Information in Contemporary Society: 14th International Conference, Conference 2019, Washington, DC, USA, March 31–April 3, 2019, Proceedings 14, pp. 590-598. Springer International Publishing, 2019.} . 

I think that despite the definition of \textit{fictio iuris} this instrument really can be used to clear the field from the responsibility gap and black box effect, by describing effectively what happens in a causal series which involves AI.
A human action, extended by an algorithm,  which creates an output.

\newpage

\bibliographystyle{siam}  
\bibliography{references}

\end{document}